\pdfoutput=1
\documentclass[11pt]{article}

\usepackage{EMNLP2023}

\usepackage{times}
\usepackage{latexsym}

\usepackage[T1]{fontenc}
\usepackage[utf8]{inputenc}

\usepackage{microtype}

\usepackage{inconsolata}

\usepackage{amsmath}

\usepackage{multirow}
\usepackage{booktabs}
\usepackage{arydshln}  
\usepackage{graphicx}
\usepackage{listings}  
\usepackage{xcolor}

\newcommand\blfootnote[1]{%
  \begingroup
  \renewcommand\thefootnote{}\footnote{#1}%
  \addtocounter{footnote}{-1}%
  \endgroup
}

\title{Tuna: Instruction Tuning using Feedback from Large Language Models}
\author{$\textbf{Haoran Li}^{\bf 1, \dagger}, \textbf{Yiran Liu}^{\bf 3,\ddagger}, \textbf{Xingxing Zhang}^{\bf 2}, \textbf{Wei Lu}^{\bf 1}, \textbf{Furu Wei}^{\bf 2}$ \\
  $^1$StatNLP Research Group, Singapore University of Technology and Design \\
  $^2$Microsoft Research Asia, $^3$Tsinghua University \\
  \texttt{haoran2\_li@mymail.sutd.edu.sg, wei\_lu@sutd.edu.sg} \\
  \texttt{liu-yr21@mails.tsinghua.edu.cn}, \texttt{\{xizhang,fuwei\}}@microsoft.com
  }

\begin{document}
\maketitle

\begin{abstract}
Instruction tuning of open-source large language models (LLMs) like LLaMA, using direct outputs from more powerful LLMs such as Instruct-GPT and GPT-4, has proven to be a cost-effective way to align model behaviors with human preferences.
However, the instruction-tuned model has only seen one response per instruction, lacking the knowledge of potentially better responses.
In this paper, we propose finetuning an instruction-tuned LLM using our novel \textit{probabilistic ranking} and \textit{contextual ranking} approaches to increase the likelihood of generating better responses.
Probabilistic ranking enables the instruction-tuned model to inherit the relative rankings of high-quality and low-quality responses from the teacher LLM.
On the other hand, learning with contextual ranking allows the model to refine its own response distribution using the contextual understanding ability of stronger LLMs.
Furthermore, we apply probabilistic ranking and contextual ranking sequentially to the instruction-tuned LLM.
The resulting model, which we call \textbf{Tuna}, consistently improves the performance on Super Natural Instructions (119 test tasks), LMentry (25 test tasks), Vicuna QA, and can even obtain better results than several strong reinforcement learning baselines. Our code and data are available at \url{ https://github.com/microsoft/LMOps}.
\blfootnote{$\dagger, \ddagger$ Work done during internship at MSRA.}

% \addtocounter{footnote}{-1}  
% \stepcounter{footnote}\footnotetext{This is an unnumbered footnote.}  
% \footnotetext[0]{This is an unnumbered footnote.}  

% \footnote{$\dagger, \ddagger$ Work done during internship at MSRA.}

\end{abstract}

\section{Introduction}
Large language models (LLMs) have made significant progress by scaling up model size and data size \cite{elmo,bert,gpt2,gpt3,gpt4} for unsupervised pre-training and subsequently applying reinforcement learning from human feedback (RLHF) to align model responses with human preferences \cite{rlhf2017,rlhf2022}.
% In RLHF, a base LLM first needs to be finetuned with human-labeled responses to input instructions in a supervised manner.
% Then, a distinct reward model needs to be trained with human-annotated comparison data.
% Lastly, the proximal policy optimization (PPO) algorithm \cite{ppo} is applied on both the LLM and the reward model to better align with human values.
% Despite its effectiveness, the RLHF approach has certain limitations, such as the necessity for a large amount of human annotations and a costly, time-consuming data collection process.
More recently, instruction tuning \cite{wei2022finetuned} with Self-Instruct algorithm \cite{self_instruct} has emerged as a cost-effective method for aligning with human preferences. 
In this approach, open LLMs like LLaMA \cite{llama} can be finetuned on instruction-following data generated by OpenAI GPT using the Self-Instruct algorithm.
The Alpaca model \cite{alpaca} exemplifies this technique, which enables close alignment with human preferences while reducing dependence on human-labeled data.

\begin{figure}[t]
    \centering
    \includegraphics[scale=0.5]{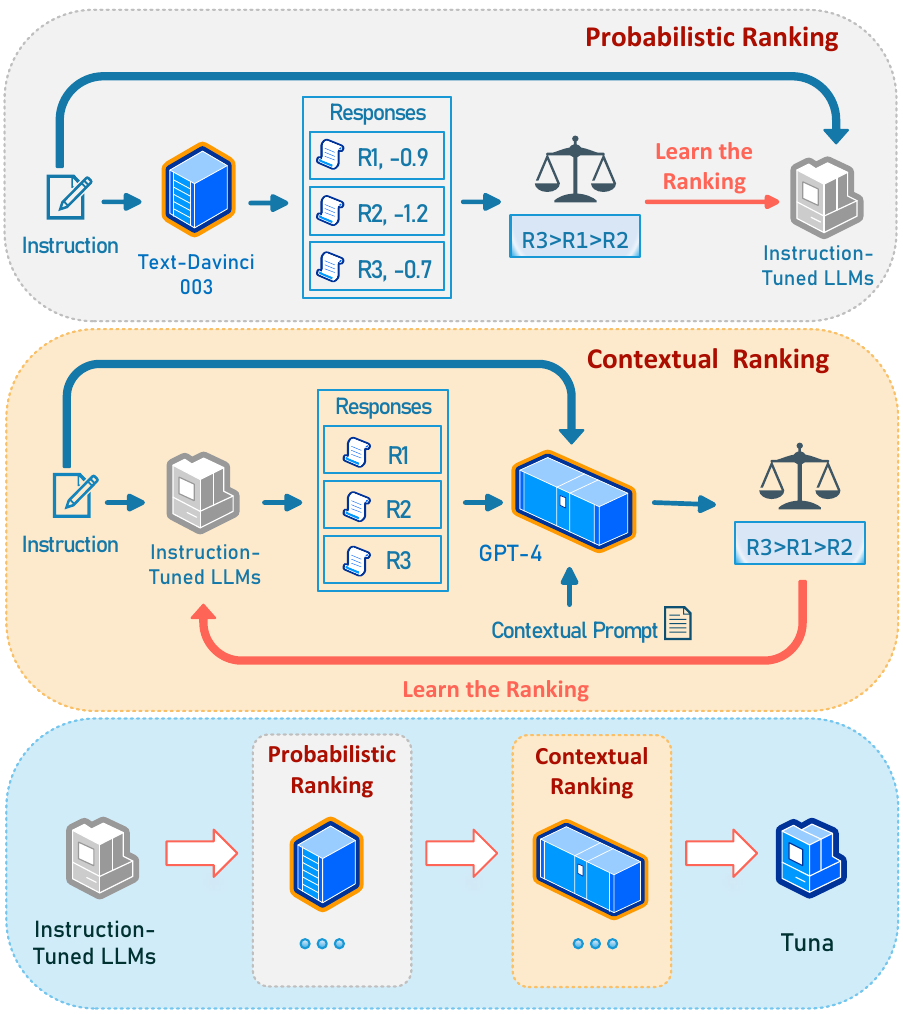}
    \caption{The finetuning process using probabilistic ranking (top), contextual ranking (middle), and a combination of both (bottom).}
    \label{fig1}
\end{figure}

However, instruction tuning offers only a broad guideline for the base LLMs to transition from ``next token prediction'' to a more interactive, instruction-following style. 
As a result, the model may learn some superficial features or styles from the instruction data but still lacks a deeper understanding of what constitutes a preferred response.
For instance, when given a question like ``Give three tips for staying healthy'', a base LLM may generate fluent yet undesirable continuations, while an instruction-tuned LLM could offer three general tips.
Humans might prefer more detailed tips over general tips, but such tips are less likely to be sampled since they have lower likelihood within the current model distribution.
This can be attributed to the fact that they are either unseen during instruction tuning or hard to be sampled due to the exposure bias \cite{mixer}.
% For instance, when given a question like ``Give three tips for staying healthy'', a base LLM may generate repetitive continuations or even empty responses, aside from the end-of-sequence token, due to its basic next-token-prediction capability.
% With instruction-tuning, the LLM is exposed to desired response patterns repeatedly, causing the model's distribution given the instruction to gradually shift toward these general patterns.
% However, at this stage, the model's instruction-following ability has only been elicited initially since it has been exposed to just one response per instruction.
% Consequently, it possesses little to no knowledge of how to produce an even better response.

To address this, we propose further finetuning of an instruction-tuned LLM to discern the quality of multiple responses more precisely, using our novel probabilistic ranking (Sec. \ref{sec:prob_rank}; Fig. \ref{fig1} top) and contextual ranking (Sec. \ref{sec:cont_rank}; Fig. \ref{fig1} middle) approaches.
% Learning with probabilistic ranking enables the instruction-tuned LLM to model a broader range of the response distribution given the instruction, and effectively differentiate between high-quality and low-quality responses.
Probabilistic ranking enables the instruction-tuned LLM to inherit the high-quality and low-quality responses as well as their relative rankings from the teacher LLM (e.g., text-davinci-003).
In contrast, contextual ranking aims to re-balance the instruction-tuned model's own response distribution with the help of stronger LLMs (e.g., GPT-4), mitigating the exposure bias issue.

We apply probabilistic ranking and contextual ranking sequentially to an instruction-tuned model,  i.e., Alpaca \cite{alpaca}, resulting in a model called \textbf{Tuna} (Sec. \ref{sec:combine}; Fig. \ref{fig1} bottom).
We evaluate Tuna on various benchmarks, including Super Natural Instructions \cite{Wang2022SuperNaturalInstructionsGV}, which contains 119 diverse test tasks; LMentry \cite{Efrat2022LMentryAL}, comprising 25 tasks to assess the basic capabilities and robustness of LLMs; and Vicuna QA \cite{FastChat} which evaluates the model's ability to answer a diverse set of questions with the assistance of GPT-4.
Experimental results demonstrate that the Tuna model not only consistently outperforms the standard instruction-tuned models on all benchmarks, but also surpasses several strong RLHF baselines \cite{rlhf2022}.

To summarize, our contributions are as follows:
\begin{itemize}
    \item We propose \textit{probabilistic ranking} and \textit{contextual ranking}, which enable the instruction-tuned model to distinguish high-quality and low-quality responses and assign higher probability to the former accordingly.
    % generate more preferable outputs.
    \item The \textbf{Tuna} model, obtained by sequentially applying probabilistic ranking and contextual ranking on an instruction-tuned LLM, achieves better results than several strong benchmarks, including RLHF models;
    \item Our model, data and code will be released to facilitate future research.
\end{itemize}

\section{Methodology}
In this section, we describe how to obtain our \textbf{Tuna} model using the feedback from LLMs. 
We first describe the vanilla instruction tuning.
We then introduce our probabilistic ranking and contextual ranking approaches.
Lastly, we describe how to integrate both ranking approaches.

\subsection{Instruction Tuning}
\label{sec:inst_tune}
LLMs like GPT-3 \cite{gpt3} have been trained on a massive text corpus using maximum likelihood estimation (MLE):
\begin{equation}
    L_{\text{MLE}}(y) = -\frac{1}{|y|} \sum_t \log p(y_t|y_{<t};\theta),
\end{equation}
where $\theta$ represents the parameters of the base model.
The pre-training objective function compels the model to predict the next token $y_t$ given its prefix $y_{<t}=[y_0, y_1,...,y_{t-1}]$.
A sufficiently-trained LLM can generate fluent continuations given almost any prefix.
% , demonstrating its strong generative capability.
% However, the generated continuations may not always be meaningful or desirable to humans.
However, the generated continuations may not align well with human preferences.
% Since the ultimate goal of an LLM is to meet human demands, it is crucial to encourage the model to produce human-preferred content in response to instructions.
As the primary goal of an LLM is to assist humans, it becomes essential to encourage the generation of content that follows human instructions and aligns with human preferences.
The current dominant approach to enhance LLMs' instruction-following ability is called \textit{instruction tuning} \cite{Mishra2021CrossTaskGV, wei2022finetuned, alpaca}, which finetunes the base LLMs in a supervised manner on instruction-response pairs $\{i,r\}$ (where $i$ is an instruction and $r$ is its response) using MLE:
\begin{equation}
\label{eq2}
    L_{\text{MLE}}(i, r) = -\frac{1}{|r|}  \log p(r|i; \theta'),
\end{equation}
where $\theta'$ represents the parameters of the instruction-tuned model.
After instruction tuning, we expect the model distribution $p(\cdot|i;\theta')$ to allocate higher probabilities to proper responses like $r$ rather than undesirable continuations.

Note that the responses in instruction-response pairs can either be annotated by humans\footnote{\url{https://huggingface.co/datasets/databricks/databricks-dolly-15k}} or generated by strong LLMs, such as Instruct-GPT or GPT-4 \cite{self_instruct}.
A prevalent and cost-effective approach for generating instruction tuning data is the Self-Instruct algorithm \cite{self_instruct}.
Specifically, it uses a strong LLM, e.g., text-davinci-003, to create instructions based on a few seed instructions, and then generates a single response for each instruction using the same LLM.

\subsection{Probabilistic Ranking} 
\label{sec:prob_rank}
% The first approach is to query the proprietary LLMs, such as \texttt{text-davinci-003}, to generate $N$ new responses for each instruction (see Fig. \ref{fig1} top).
Instruction tuning with the data generated by the Self-Instruct algorithm is essentially a form of sequence-level distillation \cite{seq_dist}.
The rationale behind this class of distillation method is that the current commercial LLMs have significantly better capabilities than their open-source counterparts.
Instead of learning from the single-response data, our \textit{probabilistic ranking} approach leverages the relative rankings of multiple responses based on the teacher model's probabilities for better pseudo label distillation (see Fig. \ref{fig1} top).

Let $r$ denote the original response for instruction $i$ in the instruction tuning dataset. 
We query strong (teacher) LLMs, such as \texttt{text-davinci-003}, to generate $N$ new responses for $i$. 
Let $r^{(0)}, r^{(1)}, \dots, r^{(N-1)}$ denote these new responses, and $p(r^{(0)} | i), p(r^{(1)} | i), \dots, p(r^{(N-1)} | i)$ denote their probabilities. 
While the teacher LLMs are expected to produce responses of comparable quality on average, there will inevitably be some variation in the quality of the generated responses.
This inherent variability manifests itself in various aspects, such as differences in accuracy \cite{selfconsistency}, response length, and level of details provided \cite{howfar}.

Intuitively, if a model is perfectly distilled, the relative probabilities assigned to two samples should be the same as those of the teacher model. 
Specifically, let $p(r^{(j)} | i; \theta')$ and $p(r^{(k)} | i; \theta')$ denote the probabilities of $r^{(j)}$ and $r^{(k)}$ w.r.t. the student model. 
If $p(r^{(j)} | i) > p(r^{(k)} | i)$, then $p(r^{(j)} | i;\theta') > p(r^{(k)} | i;\theta')$. 
We use the following normalized log-likelihood as the teacher model quality score to account for differences in response lengths:
\begin{equation}
\label{eq:logp}
    s(i, r^{(k)}) = \frac{\log p(r^{(k)} | i)}{|r^{(k)}|^{\beta}}, \quad k=\{0,...,N-1\}
\end{equation}
% To capture the differences in generation quality, we calculate the scaled log-likelihood as a proxy measure.
% We denote the log-likelihood of the $k$-th response as $\log p\big(r^{(k)}\big)$.
% Next, we introduce a scaling factor to the log-likelihood as follows:
% \begin{equation}
% \label{eq3}
%     s(r^{(k)}) = \frac{\log p(r^{(k)})}{|r^{(k)}|^{\beta}}, \quad k=\{0,...,N-1\},
% \end{equation}
where $|r^{(k)}|$ is the length of $r^{(k)}$ and $\beta$ represents the length penalty.
%{\color{red} Similarly, we denote the quality score of the instruction-tuned model as $s_{\theta'}(i, r^{(k)})$.}

% This scaled log-likelihood $s(r^{(k)})$ serves as an indicator of the quality of generated responses, as it accounts for both the model's confidence in generating a specific response and the influence of response length on the likelihood.
% By considering these factors, the scaled log-likelihood provides a meaningful and fair assessment of the quality of the generated responses.

We then rank those responses in decreasing order based on $s(i, r^{(k)})$.
% The resulting dataset will be $\{i, r, (r^{[0]},... r^{[N-1]})\}$, where $i,r$ are from the original instruction tuning data, and $r^{[j]}$ will be considered to have a better quality than $r^{[k]}$, if $j<k$.
The resulting instruction-response pairs become $\{i, r, (r^{[0]},... r^{[N-1]})\}$, where $i, r$ are from the original instruction tuning data, and $r^{[j]}$ is considered to have better quality than $r^{[k]}$, if $j<k$. 
Once we obtain the ranked responses, we can encourage our model to learn from these rankings using a pairwise ranking objective, which has been successfully employed in previous work \cite{extrasum, brio, moca, calibrate}.
The ranking objective function is as follows:
\begin{equation}
\begin{aligned}  
    & L_{\text{rank}} = \sum_{0\leq j<k \leq N-1} L_{\text{rank}}^{j,k} 
\end{aligned}
\end{equation}
\begin{equation}
    \begin{aligned}
            & L_{\text{rank}}^{j,k} = \max\Big(0, v_{\theta'}^k - v_{\theta'}^j + m\times(k-j) \Big), j<k \\
    \end{aligned}
\end{equation}
where $v_{\theta'}^{k}=\frac{1}{|r^{[k]}|} \log p\big(r^{[k]}|i;\theta' \big)$, $m>0$ is the margin hyper-parameter.
The ranking loss, $L_{\text{rank}}$, aims to teach the model to distinguish good responses from bad ones based on the teacher LLM's perspective.
In addition to $L_{\text{rank}}$, we also apply a cross-entropy loss on the original response as regularization:
\begin{equation}
\label{eq:loss}
    L = L_{\text{rank}} + \lambda L_{\text{MLE}}, \quad L_{\text{MLE}} = \frac{1}{|r|} \log p(r|i;\theta')
\end{equation}
where $r$ is the original response, and $\lambda>0$ controls the importance of $L_{\text{MLE}}$, which helps prevent over-optimization of the ranking loss.

After learning with probabilistic ranking, the model can better assign probabilities to superior and inferior responses.

\subsection{Contextual Ranking} 
\label{sec:cont_rank}
During the instruction tuning or the probabilistic ranking stage, the model is finetuned to generate a good $r$ given an instruction $i$.
However, given the same $i$ during inference, the model may still generate a relatively low-quality response $r'$.
This is related to the exposure bias problem \cite{mixer}, where the model fails to generate $r$ due to accumulated errors during the auto-regressive generation process.
% However, there may be a better response $r'$ which humans prefer but is less likely to be sampled due to its lower likelihood assigned by the instruction-tuned model, i.e., $p(r'|i;\theta')<p(r|i;\theta')$.
To address this issue, we use our \textit{contextual ranking} approach to refine the distribution of responses generated by the model itself, assigning higher probabilities to better responses with the help of strong LLMs (Fig. \ref{fig1} middle), thus alleviating exposure bias \cite{mixer}.

For each instruction, we first sample $N$ responses from the instruction-tuned model itself, i.e., $r^{(0)},r^{(1)},...,r^{(N-1)} \sim p(\cdot|i; \theta')$.
% Our second approach involves sampling responses from the instruction-tuned model $\theta'$ itself, i.e., $r \sim p(\cdot|i; \theta')$ (see Fig. \ref{fig1} middle).
% Since the model $\theta'$ has been exposed to only one response per instruction during finetuning, it tends to generate responses with varying qualities during sampling.
We hope the samples to be diverse enough so that better responses are more likely to appear in the sampled results.
To ensure diversity, we impose a constraint on the ROUGE-L \cite{Lin2004ROUGEAP} score between each pair of responses, requiring it to be less than a threshold $\tau$.
If the ROUGE-L score exceeds $\tau$, we increase the sampling temperature and resample another response.
If multiple trials still result in a ROUGE-L score above $\tau$, we retain the least similar response from the trials.
% We sample $N$ responses for each instruction and make use of the contextual understanding ability of proprietary LLMs, such as GPT-4 \cite{gpt4}, to rank them based on various aspects (see Appendix \ref{appendix_prompt}).
After obtaining $N$ responses, we leverage the contextual understanding ability of commercial LLMs, such as GPT-4 \cite{gpt4}, to rank them based on various aspects.
The ranking process consists of multiple steps.
First, we ask GPT-4 to assess whether the instruction requires an open-ended answer (e.g., story generation) or a close-ended answer (e.g., solving a math problem).
We then request GPT-4 to generate its own response as a reference.
Next, GPT-4 compares the reference response with the $N$ responses from different aspects and assign scores to each response.
For open-ended instructions, GPT-4 evaluates relevance (score 0-5), level of details$/$justification (score 0-5), and accuracy (score 0-5) of the model responses compared to its reference response.
For close-ended instructions, the evaluation criteria are accuracy (score 0-5), level of details$/$justification (score 0-5), and clarity (score 0-5).
Finally, GPT-4 ranks responses in decreasing order based on the sum of their scores (see Appendix \ref{appendix_prompt} for our complete prompt).
\textcolor{black}{We also manually evaluated GPT-4 rankings, which have achieved a strong correlation with human judgements (see Appendix \ref{appendix_gpt4_ranking}, \ref{appendix_no_pairwise_ranking}).}
% Duplicate responses will not appear in the ranking.

As in Sec. \ref{sec:prob_rank}, the resulting instruction tuning dataset becomes $\{i, r, (r^{[0]},... r^{[N-1]})\}$.
Note that the $r^{[k]},0\leq k \leq N-1$, is derived from the instruction-tuned model itself.
Lastly, we use the same objective function as in Eq. \ref{eq:loss} to encourage the model to assign higher probabilities to better responses.

% \subsection{Learn the Ranking}
% \label{sec3.3}
% After applying contextual and probabilistic ranking, the final dataset will be $\{i, r, (r^{[0]},... r^{[N-1]})\}$, where $i, r$ are from the original instruction tuning data and $r^{[j]}$ will have a better quality than $r^{[k]}$, if $j<k$.
% To learn the ranking, we employ pairwise comparisons, which have been successfully used in various works \cite{extrasum, brio, moca, calibrate}.
% Specifically, the objective function is:
% \begin{equation}
% \begin{aligned}  
%     & L_{\text{rank}} = \sum_{0\leq j<k \leq N-1} L_{\text{rank}}^{j,k} 
% \end{aligned}
% \end{equation}
% \begin{equation}
%     \begin{aligned}
%             & L_{\text{rank}}^{j,k} = \max\Big(0, s^k-s^j + m*(k-j) \Big), j<k \\
%     \end{aligned}
% \end{equation}
% where $s^k=\frac{1}{|r^{[k]}|} \log p\big(r^{[k]}|i;\theta' \big)$, $j,k$ represent the rankings, and $m>0$ denotes the margin parameter.
% The ranking loss, $L_{\text{rank}}$, provides rich signals about the characteristics of superior and inferior responses.
% In addition to $L_{\text{rank}}$, we also add a cross entropy loss for regularization:
% \begin{equation}
% \label{eq_6}
%     L = L_{\text{rank}} + \lambda L_{\text{ce}}, \quad L_{\text{ce}} = \frac{1}{|r|} \log p(r|i;\theta'),
% \end{equation}
% where $r$ comes from the instruction dataset used in the instruction tuning stage, and $\lambda>0$ controls the importance of $L_{\text{ce}}$, which helps prevent over-optimization of the ranking loss.

\subsection{Integrating Probabilistic and Contextual Ranking}
\label{sec:combine}
Given an instruction-tuned model, there are several options for further finetuning: 1) learning with probabilistic ranking alone; 2) learning with contextual ranking alone; 3) learning with probabilistic ranking followed by contextual ranking (see Fig. \ref{fig1} bottom). 
We refer to the models finetuned with these three methods as $\textbf{Tuna}_{\textbf{p}}$, $\textbf{Tuna}_{\textbf{c}}$, and \textbf{Tuna}, respectively.

To optimally integrate both probabilistic ranking and contextual ranking techniques, it is recommended to first obtain a $\text{Tuna}_p$ model, followed by applying contextual ranking to $\text{Tuna}_p$'s response distribution, resulting in the Tuna model.
There are two reasons for this choice.
First, although it is beneficial to learn the ranking of different responses from the teacher LLM's perspective (probabilistic ranking), the model might not fully capture the teacher's ranking knowledge due to its limited capacity.
Second, contextual ranking enables the model to better adapt to its own capacity by working with the model's own generations.
By generating its own responses, the model can finetune its understanding with the help of stronger LLMs and more effectively produce responses that are both closer to human preferences and compatible with its capacity constraints, alleviating the exposure bias issue \cite{mixer}.
% This is because probabilistic ranking data provides insights into the commercial LLM's distribution of responses given the instruction, which allows for a more thorough comprehension of response variability.
% In contrast, contextual ranking data is primarily concerned with optimizing the current model's existing outputs and may result in a narrower scope of response distribution, due to the model's limited capacity (compared to proprietary LLMs) and exposure bias \cite{mixer}.

\begin{table*}[h!]
    \centering
    \begin{tabular}{lcccccc}
    \toprule
         & \multicolumn{2}{c}{\textbf{Super NI}} & \multicolumn{1}{c}{\textbf{LMentry}} & \multicolumn{3}{c}{\textbf{Vicuna QA}} \\
         & \textbf{0-shot} & \textbf{2-shot} & \textbf{LMentry Score}  & \textbf{Win} & \textbf{Lose} & \textbf{Tie} \\
         \midrule
       LLaMA & 11.0 & 23.6 & 26.3  & 4\% & 92\% & 4\% \\ %3 & 74 & 3 \\
       T5-LM 11B & - & 30.2 & 20.6 & - & - & - \\
       T0 11B  & - & 32.3 & 31.6 & - & - & -  \\
       InstructGPT 175B  & - & 52.1 & 48.4 & - & - & - \\
       \midrule
       Alpaca & 36.0 & 44.5 & 31.4 & - & - & - \\
       \quad+ PPO-sim & 31.9 (-4.1) & 37.5 (-7.0) & 27.8 (-3.6) & 79\% & 16\% & 5\% \\ %63 & 13 & 4 \\
       \quad + PPO-sim-GPT4-20K  & 37.1 (+1.1) & 44.9 (+0.4) & 27.8 (-3.6) & 74\% & 22\% & 4\% \\ %59 & 18 & 3 \\
       % GPT-4 Alpaca & 32.0 & 37.8 & \textbf{33.1}  & \textbf{68} & 8 & 4 \\ 
       $\text{Tuna}_p$ & \textbf{39.4 (+3.4)} & 43.9 (-0.6) & \textbf{35.0 (+3.6)} &68\% & 27\% & 5\% \\%54 & 22 & 4 \\
       $\text{Tuna}_c$ & 37.7 (+1.7) & \textbf{46.6 (+2.1)} & 32.2 (+0.8)  & 74\%& 20\%& 6\% \\  %59 & 16 & 5 \\
       % $\text{Tuna}_c$ (Rk-3B-39K) & 35.6 (-0.4) & 40.4 (-4.1) & 33.4 (+2.0) & \textbf{79\%} & \textbf{15\%} & \textbf{6\%} \\ %\textbf{63} & \textbf{12} & \textbf{5} \\
       % $\text{Tuna}_c$ (Rk-7B-39K) & 33.5 (-2.5) & 40.3 (-4.2) & 32.5 (+1.1) & 73\% & 20\% & 7\% \\ %58 & 16 & 6 \\
       $\text{Tuna}_c$ (PRM) & 34.2 (-1.8) & 40.1 (-4.4) & 32.2 (+0.8) &  75\% & 19\% & 6\% \\%60 & 15 & 5 \\
       % $\text{Tuna}_c$ (Rk-7B-52K) & 34.6 (-1.4) & 41.1 (-3.4) & 32.0 (+0.6) & 73\% & 20\% & 7\% \\ %58 & 16 & 6 \\
       Tuna & \textbf{38.7 (+2.7)} & \textbf{45.0 (+0.5)} & \textbf{34.7 (+3.3)} & \textbf{86\%} & \textbf{10\%} & \textbf{4\%} \\ %\textbf{69} & \textbf{8} & \textbf{3} \\
       \bottomrule
    \end{tabular}
    \caption{Performance comparison of different models on Super NI, LMentry and Vicuna QA. The numbers in bold indicate the top-2 results. The numbers in parentheses indicate the performance differences compared to Alpaca.
    The results of T5-LM 11B \cite{2020t5}, T0-11B \cite{promptsource}, InstructGPT 175B \cite{rlhf2022} are taken from \citet{Wang2022SuperNaturalInstructionsGV, Efrat2022LMentryAL}. 
    The RLHF baselines PPO-sim and PPO-sim-GPT4-20K, which apply the PPO algorithm \cite{ppo}, are taken from \citet{alpacafarm}.}
    \label{table_1}
\end{table*}

\begin{table*}[h]
    \centering
    \begin{tabular}{cccccc}
    \toprule
        \textbf{Model} & Alpaca & Alpaca+PPO-sim& $\text{Tuna}_p$ & $\text{Tuna}_c$ & Tuna  \\
        \midrule
       \textbf{Score} & 2.13 & 2.95$^*$ & 2.98$^*$ & 3.15$^*$ & 3.80$^{*\dagger}$ \\
    \bottomrule
    \end{tabular}
    \caption{Human evaluation on Vicuna QA. * denotes that the model is significantly ($p<0.01$) better than Alpaca, while $\dagger$ denotes that Tuna is significantly ($p<0.01$) better than other models.}
    \label{table_human_eval_qa}
\end{table*}

\section{Experiments}
\subsection{Model and Data}
\label{sec4_1}
In our experiments, we use a 7B LLaMA model \cite{llama} as the base model.
The instruction tuning data is sourced from Alpaca \cite{alpaca}, which consists of 52K instructions paired with responses that are generated by \texttt{text-davinci-003} using the Self-Instruct algorithm \cite{self_instruct}.
We perform instruction tuning on 52K Alpaca data using recommended hyperparameters, such as a learning rate of 2e-5 and the AdamW optimizer $(0.9, 0.999)$ \cite{adamw}.\footnote{\url{https://github.com/AetherCortex/Llama-X}}
For simplicity, we also refer to the instruction-tuned model as \textbf{Alpaca}.

For probabilistic ranking, we input 52K instructions from Alpaca dataset into \texttt{text-davinci-003} to produce $N=4$ responses per instruction along with their log-likelihoods\footnote{GPT-4 is more powerful but it does not return log-likelihoods.}, with an inference temperature of 1.
We calculate response scores using Eq. \ref{eq:logp} with $\beta$ being 1.3, and rank the responses accordingly.
%In our preliminary experiments, we found that the length penalty $\beta=1.3$ is able to induce detailed responses and validated this choice on LIMA \cite{Zhou2023LIMALI} dataset (see Appendix \ref{appendix_lenpen})???????????.
Subsequently, we finetune the Alpaca model for 1 epoch with a learning rate 1e-5, margin $m=0.1$, and cross entropy regularizer weight $\lambda=1.0$.
We denote the model trained exclusively with probabilistic ranking as $\textbf{Tuna}_{\textbf{p}}$.

For contextual ranking, we sample $N=4$ responses from the Alpaca model with temperature $T=1$ for each instruction.
% We did not use smaller temperatures to avoid similar generations.
% We restrict the 4 responses to have a pairwise ROUGE-L less than 0.8, otherwise we would remove the similar response, increase the temperature by 0.1 and re-sample.
% If we cannot sample a unique enough response after 3 trials, we simply keep the least similar one.
To avoid similar generations, we ensure the pairwise ROUGE-L \cite{Lin2004ROUGEAP} between responses is less than $\tau=0.8$. 
Otherwise, we remove the similar response, increase the temperature by 0.1, and resample. 
If three trials fail to produce unique enough responses, we keep the least similar one.
We then employ GPT-4 to rank responses for the first \textbf{13K} instruction data with the GPT-4 inference temperature to be 0.
The contextual ranking prompt is shown in Table \ref{appendix_table_prompt}.\footnote{The cost of calling OpenAI API is listed in Appendix \ref{appendix_pricing}.}
The finetuning hyperprameters follow those of probabilistic ranking.
We refer to the model trained on 13K contextual ranking data of the Alpaca model as $\textbf{Tuna}_{\textbf{c}}$.

% Furthermore, we use the 13K GPT-4 ranking data to train two proxy ranking models based on StableLM-3B\footnote{\url{https://github.com/Stability-AI/StableLM}} and LLaMA-7B, which we refer to as Rk-3B and Rk-7B.
% These two ranking models are employed to re-rank Alpaca's responses on 52K instructions.
% We denote the Alpaca model trained with 13K GPT-4 contextual ranking data plus the last 39K data generated by the ranking models as $\text{Tuna}_c$ (Rk-3B-39K) and $\text{Tuna}_c$ (Rk-7B-39K). 
% The Alpaca model trained with 52K data totally generated by the ranking models are referred to as $\text{Tuna}_c$ (Rk-3B-52K) and $\text{Tuna}_c$ (Rk-7B-52K).
Furthermore, we use the 13K GPT-4 ranking data to train a proxy ranking model (PRM) based on StableLM-3B.\footnote{\url{https://github.com/Stability-AI/StableLM}}
The PRM is employed to re-rank Alpaca's responses on 52K instructions.
We refer to the Alpaca model trained with 52K ranking data totally generated by the PRM as \textbf{$\text{Tuna}_c$ (PRM)}.

Lastly, we also collect 13K GPT-4 contextual ranking data based on $\text{Tuna}_p$'s responses instead of Alpaca's. 
We refer to the model finetuned on $\text{Tuna}_p$ as \textbf{Tuna}.

We also included strong reinforcement learning baselines for comparison (i.e., PPO-sim and PPO-sim-GPT4-20K models from AlpacaFarm \cite{alpacafarm}).\footnote{We also trained our own RLHF model, which is not as good as the ones in AlpacaFarm. The comparison can be found in Appendix \ref{appendix_rlhf}}
% The learning rate is set to 1e-6 to avoid overfitting of the first 13K data.

\subsection{Evaluation}
\label{sec4_2}
% Our evaluation includes three benchmarks: Super Natural Instruction (Super NI) \cite{Wang2022SuperNaturalInstructionsGV}, LMentry \cite{Efrat2022LMentryAL} and Vicuna QA \cite{FastChat}, which provide a comprehensive assessment of the model's capabilities.

\paragraph{Super Natural Instruction (Super NI)}
Super NI \cite{Wang2022SuperNaturalInstructionsGV} contains 119 test tasks designed to evaluate a model's cross-task generalization ability.
It includes a variety of classification and generation tasks, such as textual entailment and title generation.
% The models are evaluated in an open-ended generation setting for both types of tasks, which is believed to be closer to realistic usage scenarios.\footnote{Option ranking can be adopted for classification tasks \cite{promptsource,wei2022finetuned}.}
We report both 0-shot and 2-shot performance, where 0-shot provides only an instruction (referred to as ``definition'' in their literature)  and 2-shot offers two additional positive examples.
The evaluation metric for all 119 tasks is \textbf{ROUGE-L} \cite{Lin2004ROUGEAP}, which is strongly correlated with human evaluation with a Pearson coefficient of 0.998 according to \citet{Wang2022SuperNaturalInstructionsGV}.
% \textcolor{red}{Another metric that we adopt is BERTScore \cite{Zhang*2020BERTScore:}, which is calculated with the help of pre-trained large language models.}
Greedy decoding is applied during inference.

\paragraph{LMentry}
LMentry \cite{Efrat2022LMentryAL} is a benchmark that primarily focuses on the accuracy and robustness aspects of LLMs' generations.
It contains 25 short tasks that are trivial to humans but challenging for LLMs.
% For example, in the task ``More letters'', the input can be ``Q: Which word has more letters, city or drink?'' and we can add some input perturbations by swapping ``city'' and ``drink''.
% Four different input perturbations can be applied to evaluate the model's robustness.
The final metric is \textbf{LMentry score}, which is calculated by multiplying its mean accuracy on 25 tasks with the robustness score.
The model will be evaluated in a 0-shot manner, and greedy decoding is applied during inference.
% The model will be evaluated in a 0-shot manner to avoid potential confounders of few-shot evaluation.

\paragraph{Vicuna QA}
Vicuna QA \cite{FastChat} comprises 80 test questions across 9 categories that measure an LLM's ability to generate relevant, detailed and accurate responses and it has been widely adopted in many works.
Instead of having a ground truth for evaluation, it conducts pairwise comparisons with the help of GPT-4 \cite{gpt4}.
It prompts GPT-4 to compare the outputs of our models to the Alpaca model.
% GPT-4 then scores each response on a scale of 1 to 10, with higher scores indicating better quality, and provides detailed reviews for both models.
We report the \texttt{win/lose/tie} rate against the Alpaca model.

\paragraph{Human Evaluation} Additionally, we conduct human evaluations on Vicuna QA.
Specifically, responses from five anonymous systems, namely Alpaca, Alpaca + PPO-sim, Tuna, $\text{Tuna}_p$, and $\text{Tuna}_c$, were randomly shuffled and presented to annotators who were then asked to rank these outputs. 
The scoring was designed such that the $i$-th ranked system receives a score of $6-i$, meaning the best-ranked system receives a score of 5, and the worst-ranked system receives a score of 1. 
Each question was annotated by two different annotators, and the score was averaged.

\begin{table*}[t!]
    \centering
    \begin{tabular}{lcccccc}
    \toprule
         & \multicolumn{2}{c}{\textbf{Super NI}} & \multicolumn{1}{c}{\textbf{LMentry}} & \multicolumn{3}{c}{\textbf{Vicuna QA}} \\
         & \textbf{0-shot} & \textbf{2-shot} & \textbf{LMentry Score}  & \textbf{Win} & \textbf{Lose} & \textbf{Tie} \\
         \midrule
         Alpaca & 36.0 & 44.5 & 31.4 & - & - & - \\
         Alpaca-Mul & 34.7 (-1.3) & \textbf{45.7 (+1.2)} & 33.9 (+2.5) & 42\% & 53\% & 5\% \\
         \midrule
       $\text{Tuna}_p$ & \textbf{39.4 (+3.4)} & 43.9 (-0.6) & \textbf{35.0 (+3.6)} & 68\% & 27\% & 5\% \\ %54 & 22 & 4 \\
       Tuna & \textbf{38.7 (+2.7)} & 45.0 (+0.5) & 34.7 (+3.3) & \textbf{86\%} & \textbf{10\%} & \textbf{4\%}\\%\textbf{69} & \textbf{8} & \textbf{3} \\
       % \midrule
       $\text{Tuna}_c$ & 37.7 (+1.7) & \textbf{46.6 (+2.1)} & 32.2 (+0.8)  & \textbf{74\%} & \textbf{20\%} & \textbf{6\%} \\%\textbf{59} & \textbf{16} & \textbf{5} \\
       $\text{Tuna}_{cp}$-13K & 35.7 (-0.3) & 44.0 (-0.5) & 33.5 (+2.1) & 58\% & 37\% & 5\% \\ %& 46 & 30 & 4 \\
       % \quad + davinci 9K(<0.5) (5e-6,1,0.1) dup & 33.8 & 42.0 & 35.6 & 29 & 44 & 7 \\
       % \hdashline
       % \quad + davinci 14K(<0.3) (5e-6,1,0.1) no dup & 34.6 & 43.7 & 34.0 & 48 & 26 & 6 \\
       % \quad + davinci 26K(<0.5) (5e-6,1,0.1) no dup & 36.2 & 42.7 & 32.7 & 46 & 29 & 5\\
       % \quad + davinci 25K(>0.3) (5e-6,1,0.1) no dup & 37.0 & 44.7 & 33.1 & 36 & 37 & 7 \\
       % \hdashline 
       % \quad + davinci 39K (5e-6,1,0.1) no dup & 37.3 & 44.2 & 32.1 & 38 & 37 & 5 \\
       $\text{Tuna}_{cp}$-39K & 34.8 (-1.2) & 43.4 (-1.1) & \textbf{35.4 (+4.0)} & 46\% & 48\% & 6\% \\ %37 & 38 & 5 \\
       % \quad + davinci 39K (5e-6,0.5,0.1) no dup & 18.3/36.1 & 24.2/41.1 & 31.5 & 49 & 25 & 6 \\
       % \hdashline
       % \quad + davinci 52K (1e-5,1,0.1) & 38.1 & 45.1 & 35.0 & 41 & 35 & 4 \\
       $\text{Tuna}_{cp}$-52K & 35.0 (-1.0) & 42.6 (-1.9) & 33.8 (+2.4) & 51\% & 41\% & 8\% \\ %41 & 33 & 6 \\
       % \quad + davinci 52K (2e-6,1,0.1) & 37.2 & 45.5 & \textbf{36.8} & 44 & 33 & 3 \\
       % \quad + davinci 52K (5e-6,0.1,0.05) & 20.5/38.1 & 25.6/42.2 & 35.3 & 53 & 24 & 3 \\
       % \quad + davinci 52K (5e-6,0.5,0.05) & 21.0/\textbf{39.5} & 29.2/46.4 & 35.4 & 44 & 32 & 4 \\
      
       \midrule
       mix-Tuna-52K & 37.7 (+1.7) & 44.2 (-0.3) & 30.0 (-1.4) & 70\% & 23\% & 7\% \\% 56 & 18 & 6  \\
       % 13K GPT-4, 39K Davinci (1e-5,0.5,0.1) & 19.3/37.0 & 27.1/44.0 & 29.9 & 59 & 16 & 5  \\
       % \midrule
       % 52K GPT-4, 52K Davinci (2e-5,1,0.1) & 33.9 & 38.2 & 30.9 & 8 & 67 & 5 \\
       % 52K GPT-4, 52K Davinci (1e-5,1,0.1) & 35.6 & 42.3 & 32.0 & 14 & 63 & 3 \\
       mix-Tuna-104K & 36.0 (+0.0) & 40.0 (-4.5) & 32.6 (+1.2) & 55\% & 40\% & 5\% \\ %44 & 32 & 4 \\
       \bottomrule
    \end{tabular}
    \caption{Different combinations of probabilistic ranking data and contextual ranking data. The numbers in bold represent the top-2 results. The numbers in parentheses represent the performance difference compared to Alpaca.}
    \label{table_2}
\end{table*}

\subsection{Main Results}
\label{subsec_main_results}
The main results are presented in Table \ref{table_1}. 
After instruction tuning, Alpaca demonstrates significant performance improvements over LLaMA on all three benchmarks. 
This highlights the successful transition from the ``next token prediction'' paradigm to a more interactive instruction-following paradigm.

Furthermore, both contextual and probabilistic ranking enhance performance across all three benchmarks.
Specifically, $\text{Tuna}_c$ exhibits more improvement on the Super NI\footnote{ROUGE is used as the default metric on Super NI. However, our results follow the same trend using BERTScore (see Appendix \ref{sec:bert_score}).} 2-shot results while $\text{Tuna}_p$ performs better on Super NI 0-shot and LMentry, narrowing the performance gap with much larger models like InstructGPT-175B.
Since the 2-shot input is longer than 0-shot, we conjecture that contextual ranking might be more beneficial for longer sequence generation than probabilistic ranking.
On the Vicuna QA benchmark, both $\text{Tuna}_p$ and $\text{Tuna}_c$ outperform Alpaca significantly on nearly $70\%$ of the questions, as evaluated by GPT-4.
Upon comparison with the RLHF baselines, $\text{Tuna}_p$ and $\text{Tuna}_c$ consistently demonstrate superior performances on both the Super NI and LMentry benchmarks. 
However, when it comes to the Vicuna QA benchmark, their performance is marginally lower than that of the RLHF baselines.
Moreover, Tuna achieves the best performance on Vicuna QA while maintaining competitive scores on Super-NI and LMentry.
Human results on Vicuna QA (see Table \ref{table_human_eval_qa}) also confirm that humans prefer the responses from our models.

Furthermore, $\text{Tuna}_c$ (PRM) demonstrates comparable performance to $\text{Tuna}_c$ on Vicuna QA and LMentry, but it underperforms both $\text{Tuna}_c$ and Alpaca on Super NI.
This suggests that although the PRM has primarily learned ranking from the GPT-4 contextual ranking data, it also introduces some noise during the learning process. 
Overall, it is more effective to learn directly from GPT-4 contextual ranking data.\footnote{Experiments with more PRMs can be found in App. \ref{appendix_prm}.}

\subsection{Ablation Study}
In this subsection, we delve deeper into the performance of our approach by examining several aspects, including: (a) the effect of more responses in instruction tuning, (b) the order of applying two ranking methods, (c) the influence of the cross entropy regularization, (d) the amount of probabilistic ranking data, and (e) the risks of GPT-4 evaluation.

% \subsection{The Effect of More Responses Per Instruction}
% \subsubsection{The Effect of More Responses Per Instruction}
\paragraph{More Responses in Instruction Tuning}
\label{subsubsec_more_resp}
We explore whether Tuna's effectiveness is solely due to the increased response data by examining the impact of adding more responses per instruction during instruction tuning. 
We create a new model, Alpaca-Mul, by adding four extra responses from the probabilistic ranking dataset to the Alpaca dataset and fine-tuning the LLaMA model using Eq. \ref{eq2}. 
The results are presented in Table \ref{table_2}.

Upon evaluation on Super NI, Alpaca-Mul's performance is nearly identical to that of Alpaca but falls short when compared to the 0-shot settings of $\text{Tuna}_p$ and Tuna. 
On LMentry, Alpaca-Mul outperforms Alpaca, yet it still does not reach the performance levels of $\text{Tuna}_p$ and Tuna. 
Interestingly, in the Vicuna QA task, Alpaca-Mul slightly underperforms compared to Alpaca.

These findings suggest that merely adding more responses without differentiating them does not necessarily lead to improved response generation. 
Overall, the results of Alpaca-Mul indicate that Tuna's superior performance cannot be solely attributed to the availability of more response data.

% \subsection{Combine Contextual and Probabilistic Ranking}
\paragraph{Integration Order}
\label{subsubsec_combine}
An alternative approach to Tuna involves first training the $\text{Tuna}_c$ model, and subsequently continuing training the $\text{Tuna}_c$ model with probabilistic ranking data. 
The resulting model is referred to as $\text{Tuna}_{cp}$.

We explore various strategies for training $\text{Tuna}_{cp}$: 1) finetuning $\text{Tuna}_{c}$ with the first 13K probabilistic ranking data ($\text{Tuna}_{cp}$-13K); 2) finetuing $\text{Tuna}_{c}$ model with last 39K probabilistic ranking data ($\text{Tuna}_{cp}$-39K); 3) finetuning $\text{Tuna}_{c}$ model with 52K probabilistic ranking data ($\text{Tuna}_{cp}$-52K).
Additionally, we also try to finetune original Alpaca model with a combination of 13K GPT-4 contextual ranking data (generated from Alpaca model's responses) and the last 39K probabilistic ranking data (mix-Tuna-52K).
We also finetune Alpaca model with 52K contextual ranking data (13K GPT-4 contextual ranking $+$ 39K ranking-model-generated data) plus 52K probabilistic ranking data (mix-Tuna-104K). 
The training details are listed in the Appendix \ref{appendix_hyper}.
The results are listed in Table \ref{table_2}.

\begin{table}[t!]
    \centering
    \begin{tabular}{lcccc}
    \toprule
        rank & 1 & 2 & 3 & 4 \\
    \midrule
        contextual ranking & 66.4 & 55.2 & 51.4 & 44.8 \\
        prob. ranking & 55.8 & 54.3 & 52.5 & 49.4 \\
        PRM & 69.2 & 57.8 & 50.9 & 44.7 \\
    \bottomrule
         
    \end{tabular}
    \caption{The average ranking lengths of contextual ranking data, probabilistic ranking data and the data generated by the proxy ranking model (PRM).}
    \label{table_3}
\end{table}

None of the combination strategies consistently outperform both $\text{Tuna}_p$ and $\text{Tuna}_c$ across the Vicuna QA and Super NI benchmarks.
On LMentry, however, finetuning $\text{Tuna}_{c}$ with probabilistic ranking data is beneficial, especially when no duplicate data is present ($\text{Tuna}_{cp}$-39K). 
This suggests that shorter probabilistic ranking data are beneficial when high accuracy and robustness are top priority.

Interestingly, $\text{Tuna}_{cp}$ is not comparable to Tuna, indicating that the order in which the model is trained with contextual and probabilistic ranking matters.
One plausible explanation is that both the original Alpaca data and the probabilistic ranking data are generated by \texttt{text-davinci-003}, while $\text{Tuna}_{c}$ has significantly shifted the model distribution by re-ranking the Alpaca model's responses, making it challenging to finetune $\text{Tuna}_{c}$ with probabilistic ranking data again.

\begin{figure}[t!]
    \centering
    \includegraphics[scale=0.55]{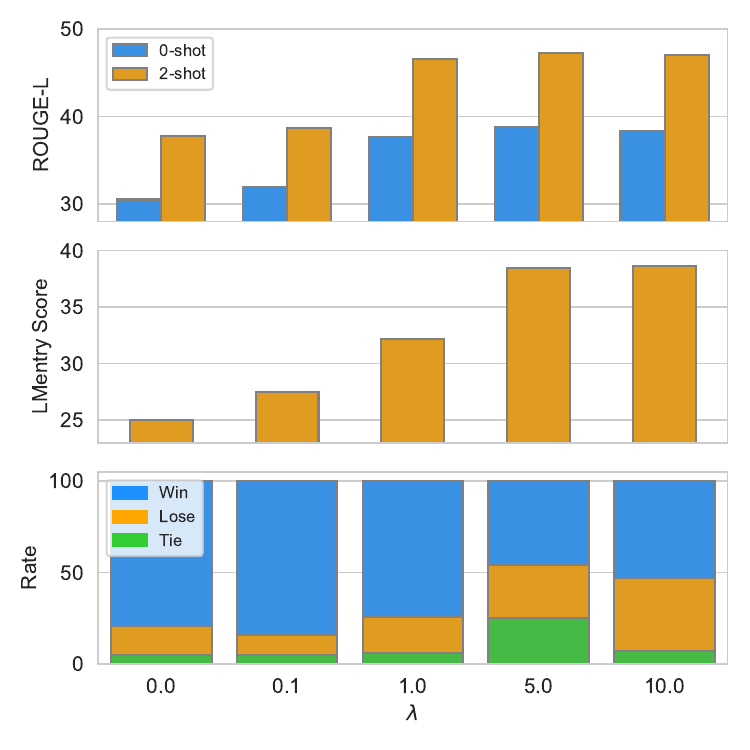}
    \caption{The effect of varying the weight $\lambda$ of cross entropy regularization in Eq. \ref{eq:loss} on $\text{Tuna}_{c}$. The win/lose/tie rate on Vicuna is computed against Alpaca.}
    \label{fig:lambda}
\end{figure}

\paragraph{The Effect of Cross Entropy Regularizer}
\label{subsubsec_lambda}
We examine the influence of the weight $\lambda$ of the cross entropy regularizer in Eq. \ref{eq:loss} on performance by varying $\lambda$ across different values: $\{0, 0.1, 1, 5, 10\}$ while training the $\text{Tuna}_{c}$ model.
Fig. \ref{fig:lambda} illustrates that as $\lambda$ increases, the performance on accuracy-oriented benchmarks such as Super NI and LMentry improves, while the performance on open questions does not necessarily follow the same trend. 
On one hand, this finding suggests that with a small $\lambda$, learning with contextual ranking may induce long and detailed answers, but those answers are not always accurate.
On the other hand, it implies that accuracy-oriented benchmarks and open QA benchmarks are complementary, and researchers should consider more diverse test cases to thoroughly evaluate a model \cite{howfar}.

\begin{figure}[t]
    \centering
    \includegraphics[scale=0.6]{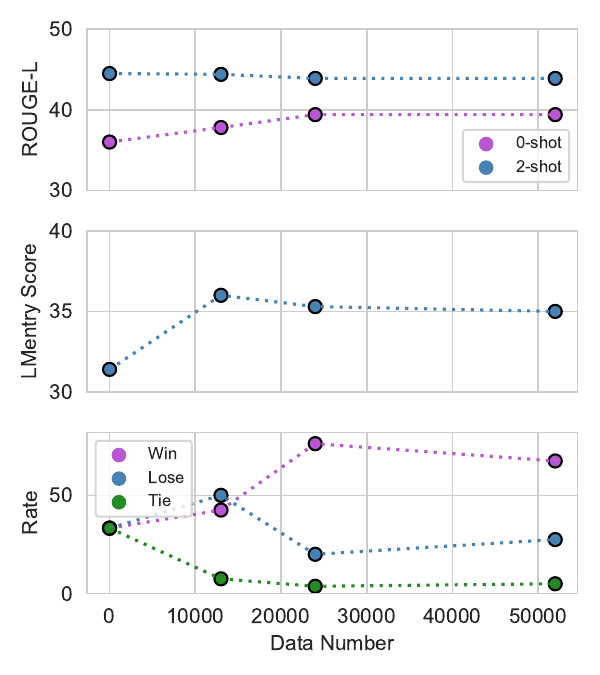}
    \caption{The effect of varying the number of probabilistic ranking data on $\text{Tuna}_p$.}
    \label{fig:data_num}
\end{figure}

\paragraph{The Amount of Probabilistic Ranking Data}
\label{subsubsec_amount_data}
We investigate the impact of varying the amount of probabilistic ranking data used for finetuning the $\text{Tuna}_{p}$ model by testing different data sizes, i.e., $\{0, 13000, 24000, 52000\}$.
$0$ refers to the Alpaca model. 
The results, shown in Fig. \ref{fig:data_num}, reveal that for probabilistic ranking, 13K data points are sufficient for Super NI and LMentry, while Vicuna QA requires 24K data points.
We conjecture that this saturation phenomenon can be attributed to two reasons.
First, 52K Alpaca instructions generated by Self-Instruct algorithm are not diverse enough, as new instructions are produced by \texttt{text-davinci-003} using prompt instructions sampled from a limited seed task pool. 
Second, instruction tuning itself may only require a limited amount of data to perform behavior cloning, as discussed in \citet{Zhou2023LIMALI}.
Thus, we can further reduce the cost of probabilistic ranking data generation by half.

% \subsection{The Risks in GPT-4 Evaluation on Open QA}
\paragraph{The Risks in GPT-4 Evaluation}
\label{subsubsec_risks}
We present evidence that evaluating a model on open QA with the help of GPT-4 may be risky.
Table \ref{table_3} displays the ranking length of our proxy ranking model (PRM). 
It shows that the PRM has inherited GPT-4 ranking's bias towards longer outputs \cite{alpaca_eval}.
However, as we discussed in Sec. \ref{subsec_main_results}, the data generated by the PRM is not as good as the original 13K contextual ranking data, as assessed by more targeted automatic evaluations like Super NI and LMentry.
Despite the inferior quality of the PRM-generated data, the performance on Vicuna QA remains almost unaffected (see $\text{Tuna}_{c}$ (PRM) in Table \ref{table_1}).
This observation suggests that evaluating LLMs on open QA with GPT-4 may not always be as accurate as it appears, echoing the findings of \citet{howfar}.
It highlights the need for more representative test questions or additional targeted benchmarks for evaluation.
% cite 2 papers that says we need better annotations.

% \section{Ablation Study}
% In this section, we perform additional analyses to investigate our finetuning objective function Eq. \ref{eq3} and the usage of probabilistic ranking data.

\section{Related Work}
\paragraph{Instruction Tuning}
Instruction tuning aims to improve the usability of base language models \cite{gpt3, 2020t5, Chowdhery2022PaLMSL} by finetuning them on instruction-response pairs in a zero-shot \cite{wei2022finetuned} or few-shot manner \cite{Mishra2021CrossTaskGV, Wang2022SuperNaturalInstructionsGV, mallen-etal-2023-trust}.
The instruction data can be sourced from off-the-shelf NLP benchmarks \cite{Mishra2021CrossTaskGV, wei2022finetuned, Wang2022SuperNaturalInstructionsGV} or generated by LLMs \cite{self_instruct,honovich2022unnatural,alpaca,Peng2023InstructionTW}.
% Instruction tuning initially aligns model behavior with human preferences.
% After that, reinforcement learning with human feedback can be applied to further unlock the LLMs' abilities \cite{rlhf2017,sum_hf,rlhf2022}.
% Recently, researchers have found that instruction tuning may only require a small amount of high quality data \cite{Zhou2023LIMALI} and the base LLMs is the key to gain better performance \cite{falsepromise}.

\paragraph{Ranking Loss}
Learning through re-ranking sequence-level outputs has been studied in sequence-to-sequence models \cite{Wiseman2016SequencetoSequenceLA, edunov-etal-2018-classical, brio, moca}.
BRIO and MoCa algorithms \cite{brio,moca} adopt a pairwise ranking loss to guide the model to generate summaries with higher ROUGE scores \cite{Lin2004ROUGEAP}.
In this paper, we use GPT-4's \cite{gpt4} strong contextual understanding ability and \texttt{text-davinci-003}'s \cite{rlhf2022} intrinsic probability measures for ranking.
In parallel with our work, \citet{Yuan2023RRHFRR} also propose pairwise ranking loss for finetuning LLMs.
Key differences include: 1) our pipeline finetuning strategy; 2) our focus on ranking the model's responses; 3) our use of the original response for cross entropy regularization, while they select the highest-reward response.
% The main differences are 1) we employ a pipeline finetuning strategy while their method is similar to our contextual ranking method; 2) our contextual ranking focuses on ranking the model's own responses while they rank responses from multiple sources; 3) they choose the response with the highest reward for cross entropy regularization while we use the original response.
Additionally, \citet{gptrank} also employs GPT models for finetuning BART \cite{BART} on the summarization task.

\paragraph{Pre-Trained Model Evaluation}
% Compared to statistical evaluation metrics such as BLEU \cite{Papineni2002BleuAM} and ROUGE \cite{Lin2004ROUGEAP}, 
Large pre-trained models are powerful evaluation metrics due to their strong contextual understanding ability, such as BERTScore \cite{Zhang*2020BERTScore:}, BARTScore \cite{bartscore}, MoverScore \cite{moverscore}, COMET \cite{comet}, and GPTScore \cite{Fu2023GPTScoreEA}.
More recently, there are more evaluation strategies based on GPT-3.5 and GPT-4 \cite{geval,humanlike}.
% The closely related work to our probabilistic ranking is GPTScore, but there are several key differences: 1) GPTScore compares the likelihood of different models' outputs under a GPT model while we care about the likelihood of the outputs sampled from the GPT model itself (i.e., text-davinci-003); 2) GPTScore explicitly requires a prompt template that defines general evaluation protocols (e.g., factuality and influency) as input whereas the input required by probabilistic ranking is an arbitrary instruction; 3) GPTScore is obtained by dividing the sequence likelihood by the sequence length while we use a scaled token-level log likelihood as the ranking score.

\section{Conclusion}
In this paper, we propose to finetune an instruction-tuned LLM using our probabilistic ranking approach ($\text{Tuna}_{p}$), contextual ranking approach ($\text{Tuna}_{c}$), and a combination of both (Tuna).
Our comprehensive experiments demonstrate consistent performance improvements across three benchmarks: Super Natural Instructions (119 test tasks), LMentry (25 test tasks), and vicuna QA.
Furthermore, our methods outperform popular reinforcement learning from human feedback baselines that rely on the proximal policy optimization algorithm.
These findings underscore the effectiveness of our approach in enhancing the performance of instruction-tuned LLMs and pave the way for future research in this area.

\section*{Limitations}
Despite the promising results achieved by our Tuna model, there are several limitations that should be acknowledged.
The first limitation is GPT-4 ranking inconsistency.
In our experiments, we relied on GPT-4 for contextual ranking, which may introduce bias due to the inconsistency in its ranking performance. 
As a powerful LLM, GPT-4 is generally expected to provide accurate and reliable rankings; however, it may still be sensitive to the phrasing or structure of prompts \cite{alpacafarm}. 
This inconsistency may lead to sub-optimal rankings and potentially affect the overall performance of the Tuna model. 
In future work, it would be beneficial to design more robust prompts that can mitigate the impact of GPT-4's ranking inconsistencies.
Another limitation is the evaluation benchmark.
In this paper, we evaluated the Tuna model on three benchmarks, which provided a diverse range of tasks and challenges. 
However, it is unclear how well the Tuna model would generalize to other types of tasks, domains, or languages. 
Further research is needed to explore the applicability of the Tuna model to a broader range of problems and settings.
The last limitation is the reliance on the use of proprietary LLMs, such as GPT-4 and \texttt{text-davinci-003}, for generating responses and rankings. 
This dependency may limit the accessibility and reproducibility of our method for researchers who do not have access to these proprietary models.
Developing alternative methods that can leverage open-source LLMs or other ranking mechanisms would be a valuable direction for future research.

% \section*{Ethics Statement}
% \section*{Ethics Statement}
% Scientific work published at EMNLP 2023 must comply with the \href{https://www.aclweb.org/portal/content/acl-code-ethics}{ACL Ethics Policy}. We encourage all authors to include an explicit ethics statement on the broader impact of the work, or other ethical considerations after the conclusion but before the references. The ethics statement will not count toward the page limit (8 pages for long, 4 pages for short papers).

\section*{Acknowledgements}
We would like to thank reviewers for their valuable feedback. This research/project is supported by Ministry of Education, Singapore, under its Tier 3 Programme (The Award No.: MOET320200004), the National Research Foundation Singapore and DSO National Laboratories under the AI Singapore Program (AISG Award No: AISG2-RP-2020-016), and Ministry of Education, Singapore, under its Academic Research Fund (AcRF) Tier 2 Programme (MOE AcRF Tier 2 Award No: MOE- T2EP20122-0011). Any opinions, findings and conclusions or recommendations expressed in this material are those of the authors and do not reflect the views of the Ministry of Education, Singapore.

% \bibliography{custom}
\bibliography{emnlp2023}
\bibliographystyle{acl_natbib}

\appendix
\section{The Length Penalty $\beta$ for Probabilistic Ranking Data}
\label{appendix_lenpen}
In our preliminary experiments, we found that the length penalty $\beta=1.3$ is able to induce detailed responses and validated this choice on LIMA \cite{Zhou2023LIMALI} dataset.
We finetune the $\beta$ parameter in Eq. \ref{eq:logp} using the LIMA training dataset, which contains 1030 high-quality expert instruction annotations, allowing LLaMA-65B to be finetuned and achieve remarkably strong performance across a wide range of topics.
Note that the training set also contains 50 modified Super NI examples but they are from the training tasks while we test our models on 119 Super NI test tasks.
Specifically, we first obtain $\text{Tuna}_{p}$ models with probabilistic ranking data scored with different $\beta$.
Then, we compute the token-level negative log-likelihood (NLL) of the output of each LIMA instance under different $\text{Tuna}_{p}$ models and average the token likelihood over the whole LIMA training set.
The results are shown in Table \ref{appendix_table_lenpen}.
It can be seen that with $\beta=1.3$, the model can achieve the best NLL on LIMA training set.
Thus, we set $\beta=1.3$ in our experiments.

\begin{table}[h!]
    \centering
    \begin{tabular}{ccccccc}
    \toprule
        $\beta$ & 0.9 & 1.0 & 1.1 & 1.2 & 1.3 & 1.4 \\
        \midrule
        NLL & 2.14 & 2.12 & 2.11 & 2.10 & \textbf{2.09} & \textbf{2.09} \\
    \bottomrule
    \end{tabular}
    \caption{The token-level log-likelihood of LIMA training set under $\text{Tuna}_p$ models trained with probabilistic ranking data scored with different $\beta$.}
    \label{appendix_table_lenpen}
\end{table}

\section{OpenAI API Pricing}
\label{appendix_pricing}
We list the cost of calling OpenAI API models in Table \ref{appendix_table_pricing}.\footnote{\url{https://openai.com/pricing}}
The human labeling cost per 1K examples is estimated based on the pricing listed in \citet{alpacafarm}, at 0.25\$ for each comparison.
For each data example, there are 4 responses and thus $4*(4-1)/2=6$ comparisons.
Thus, the human labor cost per 1K examples is 1500\$.

\begin{table*}[t!]
    \centering
    \begin{tabular}{cccc}
    \toprule
         & Data Num & Model & Price   \\
        \midrule
       Probabilistic Ranking & 52K & \texttt{text-davinci-003} & 275\$ \\
       Contextual Ranking &  13K & GPT-4-0314 & 380\$ \\
       \midrule
       Human & 1K & - &  1500\$\\
       \bottomrule
    \end{tabular}
    \caption{The estimated cost of calling OpenAI API and human labeling.}
    \label{appendix_table_pricing}
\end{table*}

\section{Training Details of $\text{Tuna}_{cp}$ and mix-Tuna}
\label{appendix_hyper}
The hyperparameters are listed in Table \ref{appendix_table_hyper}.
For models finetuned from Alpaca, i.e., $\text{Tuna}_{c}$, $\text{Tuna}_{p}$, and mix-Tuna, the learning rate is 1e-5.
The only exception is mix-Tuna-104K, whose learning rate is 5e-6 since it contains 52K duplicate data.
For models finetuned from $\text{Tuna}_{c}$ or $\text{Tuna}_{p}$, i.e., $\text{Tuna}_{cp}$ and Tuna, the learning rate is 1e-6.
We use 8 Nvidia V100-32GB GPUs for all experiments in this paper.

\begin{table*}[h!]
    \centering
    \begin{tabular}{ccccc}
    \toprule
         & LR & epoch & batch size & warmup \\
         \midrule
         
        $\text{Tuna}_{c}$ & 1e-5 & 1 & 128 & 2 \\
        $\text{Tuna}_{p}$ & 1e-5 & 1 & 128 & 2 \\
        mix-Tuna-52K & 1e-5 & 1 & 128 & 2 \\
        mix-Tuna-104K & 5e-6 & 1 & 128 & 2 \\
        Tuna & 1e-6 & 1 & 128 & 2 \\
        $\text{Tuna}_{cp}$-13K & 1e-6 & 1 & 128 & 2   \\
        $\text{Tuna}_{cp}$-39K & 1e-6 & 1 & 128 & 2  \\
        $\text{Tuna}_{cp}$-52K & 1e-6 & 1 & 128 & 2  \\

    \bottomrule
    \end{tabular}
    \caption{The hyperparameters of training different models.}
    \label{appendix_table_hyper}
\end{table*}

\section{Other Proxy Ranking Models (PRM)}
\label{appendix_prm}
Similar to the PRM introduced in Sec. \ref{sec4_1}, we use the 13K GPT-4 ranking data to train another PRM based on LLaMA-7B, which we refer to as PRM-7B.
We denote the PRM based on StableLM-3B as PRM-3B.
These two ranking models are employed to re-rank Alpaca's responses on 52K instructions.
The Alpaca model trained with 52K data totally generated by the ranking models are referred to as $\text{Tuna}_c$ (PRM-3B-52K) and $\text{Tuna}_c$ (PRM-7B-52K).
Note that $\text{Tuna}_c$ (PRM-3B-52K) is the $\text{Tuna}_c$ (PRM) listed in Table \ref{table_1}.
We denote the Alpaca model trained with 13K GPT-4 contextual ranking data plus the last 39K data generated by the ranking models as $\text{Tuna}_c$ (PRM-3B-39K) and $\text{Tuna}_c$ (PRM-7B-39K). 

The results are listed in Table \ref{appendix_table_prm}.
We can observe that models trained with ranking data generated by both PRMS do not achieve better results on Super NI compared to $\text{Tuna}_c$.
The performances of $\text{Tuna}_c$ (PRM-3/7B-39K) is close to $\text{Tuna}_c$ (PRM-3/7B-52K), implying that the ranking model have learned 13K contextual ranking data well.
Using a larger ranking model, such as 7B, does not gain better performance, which indicates that the ranking ability might not necessarily scale with the pre-training model's capacity.
In general, the best strategy is still to learn directly from GPT-4 contextual ranking data, which contains less noise.

\begin{table*}[t!]
    \centering
    \begin{tabular}{lcccccc}
    \toprule
         & \multicolumn{2}{c}{\textbf{Super NI}} & \multicolumn{1}{c}{\textbf{LMentry}} & \multicolumn{3}{c}{\textbf{Vicuna QA}} \\
         & \textbf{0-shot} & \textbf{2-shot} & \textbf{LMentry Score}  & \textbf{Win} & \textbf{Lose} & \textbf{Tie} \\
         \midrule
       Alpaca & 36.0 & 44.5 & 31.4 & - & - & - \\
       $\text{Tuna}_p$ & \textbf{39.4 (+3.4)} & 43.9 (-0.6) & \textbf{35.0 (+3.6)} &68\% & 27\% & 5\% \\%54 & 22 & 4 \\
       $\text{Tuna}_c$ & 37.7 (+1.7) & \textbf{46.6 (+2.1)} & 32.2 (+0.8)  & 74\%& 20\%& 6\% \\  %59 & 16 & 5 \\
       $\text{Tuna}_c$ (PRM-3B-39K) & 35.6 (-0.4) & 40.4 (-4.1) & 33.4 (+2.0) & \textbf{79\%} & \textbf{15\%} & \textbf{6\%} \\ %\textbf{63} & \textbf{12} & \textbf{5} \\
       $\text{Tuna}_c$ (PRM-7B-39K) & 33.5 (-2.5) & 40.3 (-4.2) & 32.5 (+1.1) & 73\% & 20\% & 7\% \\ %58 & 16 & 6 \\
       $\text{Tuna}_c$ (PRM-3B-52K) & 34.2 (-1.8) & 40.1 (-4.4) & 32.2 (+0.8) &  75\% & 19\% & 6\% \\%60 & 15 & 5 \\
       $\text{Tuna}_c$ (PRM-7B-52K) & 34.6 (-1.4) & 41.1 (-3.4) & 32.0 (+0.6) & 73\% & 20\% & 7\% \\ %58 & 16 & 6 \\
       Tuna & \textbf{38.7 (+2.7)} & \textbf{45.0 (+0.5)} & \textbf{34.7 (+3.3)} & \textbf{86\%} & \textbf{10\%} & \textbf{4\%} \\ %\textbf{69} & \textbf{8} & \textbf{3} \\
       \bottomrule
    \end{tabular}
    \caption{Performance comparison of different models. The numbers in bold indicate the top-2 results. The numbers in parentheses indicate the performance difference compared to Alpaca.}
    \label{appendix_table_prm}
\end{table*}

\section{Contextual Ranking Prompt}
\label{appendix_prompt}
We show the prompt that we use for GPT-4 contextual ranking in Table \ref{appendix_table_prompt}.

\begin{table*}[ht]  
\centering   
\begin{tabular}{p{15cm}}  
\toprule

Below is an instruction that describes a task, paired with an input that provides further context. Write a response that appropriately completes the request. \\
\\
\#\#\# Instruction:\\
\{\texttt{Instruction}\}  \\
\\
\#\#\# Input: \\
\{\texttt{Input}\} \\
\\
\#\#\# Response: \\
\\
\#\#\#Response 0: \\
\{\texttt{Response 0}\} \\
\\
\#\#\#Response 1: \\
\{\texttt{Response 1}\} \\
\\
\#\#\#Response 2: \\
\{\texttt{Response 2}\} \\
\\
\#\#\#Response 3: \\
\{\texttt{Response 3}\} \\
\\
We would like you to rate Response 0/1/2/3 in reply to the given instruction displayed above.\\
First, identify if the instruction requires open-ended or close-ended responses.\\
Second, you need to generate one high quality `\#\#\#Response 4' in answer to the instruction. It needs to have the same format as other responses and will be used as a reference later.\\
Third, identify if there are duplicate responses and keep only one of the duplicate responses for the following steps.\\ 
Fourth, compare Response 4 with Response 0/1/2/3/4 and assign each response an overall score on a scale of 0 to 15 where a higher score indicates better overall quality. 
For an open-ended instruction, please rate based on the relevance (score 0 to 5), level of details/justification: (score 0 to 5) and accuracy (score 0 to 5) of each response; for a close-ended instruction, please rate based on the accuracy (score 0 to 5), level of details/justification (score 0 to 5) and clarity (score 0 to 5) of each response. The ratings should have the format: `Response k: [\texttt{sum of the 3 individual scores you give to response k}]'.\\
Last, rank the responses in decreasing order of their overall scores. The ranking should have the format: `rank: [i, j ,k, l, m]'. If there are duplicate responses, keep only one of them in the rank, that is, the ranking may become: `rank: [i, j, k, l]', `rank: [i, j, k]' `rank: [i, j]' or even `rank: [i]'.  
\\
\bottomrule
\end{tabular}  
\caption{Contextual Ranking Prompt for GPT-4.}
\label{appendix_table_prompt}  
\end{table*}  
% \begin{figure*}[h!]
%     \centering
%     \includegraphics[scale=0.9]{emnlp2023-latex/fig_prompt_cropped.pdf}
%     \caption{The prompt used for GPT-4 contextual ranking.}
%     \label{appendix_fig_prompt}
% \end{figure*}

\section{Is the Contextual Ranking Prompt Too Long?}
In \cite{liu2023lost}, the authors found the "lost in the middle" phenomenon occurs at around 2K (20 documents * 100 tokens/document) tokens for GPT-4 (note we used GPT-4 in contextual ranking). 
We computed the average length of the prompt (including four responses and the ranking guidelines) used in GPT-4 ranking. 
The average length is 650 tokens, which is significantly shorter than 2K. Thus, the input length does not seem to be an issue in GPT-4 ranking. 
Our human experiments above also confirm that the GPT-4 ranking is closely aligned with human assessments (see Appendix \ref{appendix_gpt4_ranking}).

\section{Human Evaluation of GPT-4 Ranking}
\label{appendix_gpt4_ranking}
We conducted human evaluations of GPT-4 rankings on 50 questions used for contextual ranking. 
We asked annotators to rank the four system outputs produced by our model and we observe that the ranking quality by GPT-4 is reasonably good (the Spearman coefficient between the human rankings and GPT-4 rankings is 0.72). 
Furthermore, we also manually inspected the explanations given by GPT-4 for the ranking results. 
We found these explanations to be well-reasoned and plausible. 
Perhaps this is not surprising given the fact that several recent papers found GPT can be good evaluators in multiple NLP tasks \cite{Fu2023GPTScoreEA, humanlike}. 
We believe the ranking feedback of this level is sufficient for guiding our model for better training (our experiments also proved this).

\section{Why Not Choose Pairwise Ranking in GPT-4 Ranking}
\label{appendix_no_pairwise_ranking}
There are several reasons why ranking 4 responses together is preferred over pairwise rankings.
\begin{enumerate}
 \item API cost: Pairwise ranking for four responses requires (4 * 3) / 2 = 6 API calls, significantly increasing the total cost. Moreover, a loop (e.g., R1 > R2, R2 > R3, R3 > R1) could occur when R1/2/3 are of similar qualities, potentially requiring extra API calls for further validation.

\item The GPT-4 ranking quality is good enough, see Appendix \ref{appendix_gpt4_ranking}.
\end{enumerate}

\section{Comparison between Our RLHF Models and PPO-sim/PPO-sim-GPT4-20K}
\label{appendix_rlhf}
We compare our RLHF models and PPO-sim/PPO-sim-GPT4-20K from \citet{alpacafarm} on Vicuna QA.
The results can be found in Table \ref{appendix_table_rlhf}.
PPO-sim/PPO-sim-GPT4-20K have better responses and thus we choose to report the results of their models.

\begin{table}[t]
    \centering
    \begin{tabular}{lccc}
    \toprule
        	& Win	& Lose &	Tie \\
         \midrule
PPO-sim	 & 79\%	 &16\% &	5\% \\
PPO-sim-GPT4-20K &	74\% &	22\% &	4\%  \\
Our PPO (w/ GPT4 ranking data) &	55\% &	35\% &	10\% \\
\bottomrule
    \end{tabular}
    \caption{Vicuna QA results of our RLHF models and \cite{alpacafarm}.}
    \label{appendix_table_rlhf}
\end{table}

\section{BERTScore Results on Super NI}
\label{sec:bert_score}

By default, ROUGE is employed on Super NI. We additionally reported BERTScore (which is proven to be a better alternative for ROUGE) in Table \ref{tab_bertscore} and the results follow the same trends (also see Table \ref{table_1}).

\begin{table}[t!]
    \centering
    \begin{tabular}{lcc}
    \toprule
        \textbf{MODEL} & \textbf{0-shot} & \textbf{2-shot} \\
    \midrule
        Alpaca	& 0.638 & 0.705	\\
        PPO-sim & 0.594 & 0.652	\\
        PPO-sim-GPT4-20K & 0.641 & 0.714\\
        $\text{Tuna}_p$ & \textbf{0.667} &	0.690	\\
        $\text{Tuna}_c$ &	0.655 &	\textbf{0.727}	\\
        Tuna & 0.665 & 0.699	 \\
        \bottomrule
    \end{tabular}
    \caption{BERTScore evaluation of the model outputs in Super NI. Higher is better.}
    \label{tab_bertscore}
\end{table}

\section{Examples}
\label{appendix_examples}
We show some examples in \ref{appendix_table_examples}.

\begin{table*}[ht]  
\centering   
\begin{tabular}{p{15cm}}  
\toprule
Instruction: "Why is open-source important? Answer in 3-5 sentences" \\
\midrule
\textbf{Alpaca}:\\
Open-source software is important because it allows for collaboration and innovation. It allows developers to work together to create better software, and it allows users to access the source code and make changes to the software to meet their own needs. Open-source software also encourages transparency and allows users to see how the software works and how it can be improved. Additionally, open-source software is often more secure and reliable than proprietary software, as it has been tested and reviewed by a large number of developers. Finally, open-source software is often free or low-cost, making it accessible to everyone.
\\
\\
$\textbf{Tuna}_{\textbf{c}}$:\\
Open-source software is important because it allows for collaboration and innovation. It allows developers to work together to improve existing software and create new software, and it allows users to view and modify the source code to customize the software to their own needs. Open-source software also encourages transparency and accountability, as anyone can view and inspect the code to ensure that it is secure and reliable. Additionally, open-source software is often more cost-effective than proprietary software, as users can access the source code and modify it to suit their own needs without having to pay for a license. Finally, open-source software is often more reliable and secure than proprietary software, as it is developed and tested by a large community of developers.
\\
\\
$\textbf{Tuna}_{\textbf{p}}$:\\
Open-source software is important because it allows developers to access the source code of a program and make changes to it. This allows developers to customize the software to their needs and to fix any bugs or errors that may be present. Additionally, open-source software is often free to use and modify, which makes it more accessible to those who may not be able to afford proprietary software. Open-source software also allows developers to collaborate and share ideas, which can lead to the development of better and more efficient software. Finally, open-source software can be used as a learning tool, as it allows developers to see how the software works and how it is put together. This can help them to better understand programming and coding in general.
\\
\\
$\textbf{Tuna}$:\\ 
Open-source software is important because it allows developers to collaborate on projects, share code, and create better products. By making the source code available to the public, developers can easily collaborate on projects, allowing them to create better products and services. This collaboration also allows developers to quickly identify and fix bugs, as well as add new features and functionality. Additionally, open-source software is often more secure than proprietary software, as it can be more easily audited and reviewed by the public. Finally, open-source software is often more cost-effective than proprietary software, as it does not require licensing fees or other costs associated with closed-source software. Open-source software is an important part of the software industry, and is used by many businesses and individuals.
\\
\\
\bottomrule
\end{tabular}  
\caption{Example responses of different models.}
\label{appendix_table_examples}  
\end{table*}

\begin{table*}[ht]  
\centering   
\begin{tabular}{p{15cm}}  
\toprule
Instruction: "Write a Python function that generates a random number between 1 and 10 (inclusive)." \\
\midrule
\begin{lstlisting}[language=Python, caption=Alpaca]  
def random_number():
    return random.randint(1, 11)
\end{lstlisting}  

\begin{lstlisting}[language=Python, caption=$\text{Tuna}_c$]  
def random_number():
    return random.randint(1, 11)
\end{lstlisting}  

\begin{lstlisting}[language=Python, caption=$\text{Tuna}_p$]  
"""
Generate a random number between 1 and 10 (inclusive)
"""

import random

def random_number():
    return random.randint(1, 10)

if __name__ == '__main__':
    print(random_number())
\end{lstlisting}

\begin{lstlisting}[language=Python, caption=$\text{Tuna}$]  
"""
Generate a random number between 1 and 10 (inclusive)
"""

import random

def random_number():
    return random.randint(1, 10)

if __name__ == '__main__':
    print(random_number())
\end{lstlisting}  
\\
\\
\bottomrule
\end{tabular}  
\caption{Example responses of different models.}
\label{appendix_table_examples2}  
\end{table*}

\end{document}